\documentclass[sigconf]{acmart}
\AtBeginDocument{%
  }

\copyrightyear{2026}
\acmYear{2026}
\setcopyright{cc}
\setcctype{by}
\acmConference[MM '26]{Proceedings of the 34th ACM International Conference on Multimedia}{November 10--14, 2026}{Rio de Janeiro, Brazil}
\acmBooktitle{Proceedings of the 34th ACM International Conference on Multimedia (MM '26), November 10--14, 2026, Rio de Janeiro, Brazil}
\acmDOI{10.1145/3767308.3836079}
\acmISBN{979-8-4007-2213-4/2026/11}
\usepackage{multirow}
\usepackage{graphicx}
\usepackage{subcaption}
\usepackage{stmaryrd}
\usepackage{siunitx}

\begin{document}

\title{GRCF: Group-wise Ranking and Calibration Framework for Multimodal Sentiment Analysis}

\author{Manning Gao}
\orcid{0009-0008-4810-7428}
\affiliation{%
  \institution{South China Normal University}
  \city{Guangzhou}
  \state{Guangdong}
  \country{China}
}
\email{20232005149@m.scnu.edu.cn}

\author{Leheng Zhang}
\orcid{0009-0006-1648-4873}
\affiliation{%
  \institution{South China Normal University}
  \city{Guangzhou}
  \state{Guangdong}
  \country{China}}
\email{lehengzhang@m.scnu.edu.cn}

\author{Shiqin Han}
\orcid{0009-0005-0191-3746}
\affiliation{%
  \institution{South China Normal University}
  \city{Guangzhou}
  \state{Guangdong}
  \country{China}
}
\email{20222121019@m.scnu.edu.cn}

\author{Haifeng Hu}
\orcid{0000-0002-4884-323X}
\affiliation{%
 \institution{Sun Yat-sen University}
 \city{Guangzhou}
 \state{Guangdong}
 \country{China}}
\email{huhaif@mail.sysu.edu.cn}

\author{Yuncheng Jiang}
\orcid{0000-0002-0402-5382}
\affiliation{%
  \institution{South China Normal University}
  \city{Guangzhou}
  \state{Guangdong}
  \country{China}}
\email{ycjiang@scnu.edu.cn}

\author{Sijie Mai}
\orcid{0000-0001-9763-375X}
% \correspondingauthor
\authornote{indicates the corresponding author.}
\affiliation{%
  \institution{South China Normal University}
  \city{Guangzhou}
  \state{Guangdong}
  \country{China}}
\email{sijiemai@m.scnu.edu.cn}

% \renewcommand{\shortauthors}{Gao et al.}
\renewcommand{\shortauthors}{Manning Gao et al.}
\begin{abstract}
Most multimodal sentiment analysis algorithms focus on point-wise regression. While straightforward, it uses absolute values to represent subjective sentiment and overlooks the ordinal nature of sentiment, resulting in unstable predictions and poor correlation. Inspired by the way humans perceive sentiment, pair-wise ordinal learning frameworks emerged to address this gap, capturing relative order by learning from comparisons. However, they have two limitations: First, they assign uniform importance to all pairs, failing to adaptively focus on hard-to-rank samples. Second, they employ static ranking margins, which fail to reflect the varying semantic distances between sentiment groups. To address this, we propose a Group-wise Ranking and Calibration Framework (GRCF) that introduces the Adaptive Error-weighting and Dynamic Margin mechanism, which resolves these limitations by simultaneously preserving the relative ordinal structure and ensuring absolute score calibration. Specifically, Stage 1 introduces a Group-aware Ranking Loss to build a fine-grained ordinal structure by assigning higher optimization priority to confusable pairs. Stage 2 then employs an MAE-driven objective to align the prediction magnitudes. To validate its generalizability, we extend GRCF to classification tasks, including Multimodal Humor and Sarcasm Detection. It achieves state-of-the-art performance on regression datasets while also showing strong generalizability in classification tasks.
\end{abstract}




\maketitle

\section{Introduction}
\label{sec:intro}
Understanding human affect through multimodal signals (language, vision, and acoustics) has been a central pursuit in affective computing. Unlike categorical perception, human emotion unfolds along a continuous, ordinal spectrum of intensity~\cite{yannakakis2018ordinal,stoehr2023sentiment}. Sentiment intensity is inherently comparative. For instance, terms like ``desperate'' are universally understood to be more negative than ``sad''; yet, due to the inevitable information loss when mapping a high-dimensional emotional space to a low-dimensional scale, their exact absolute magnitudes remain subjective and ambiguous.
\begin{figure}[t]
\centering
\includegraphics[width=\linewidth]{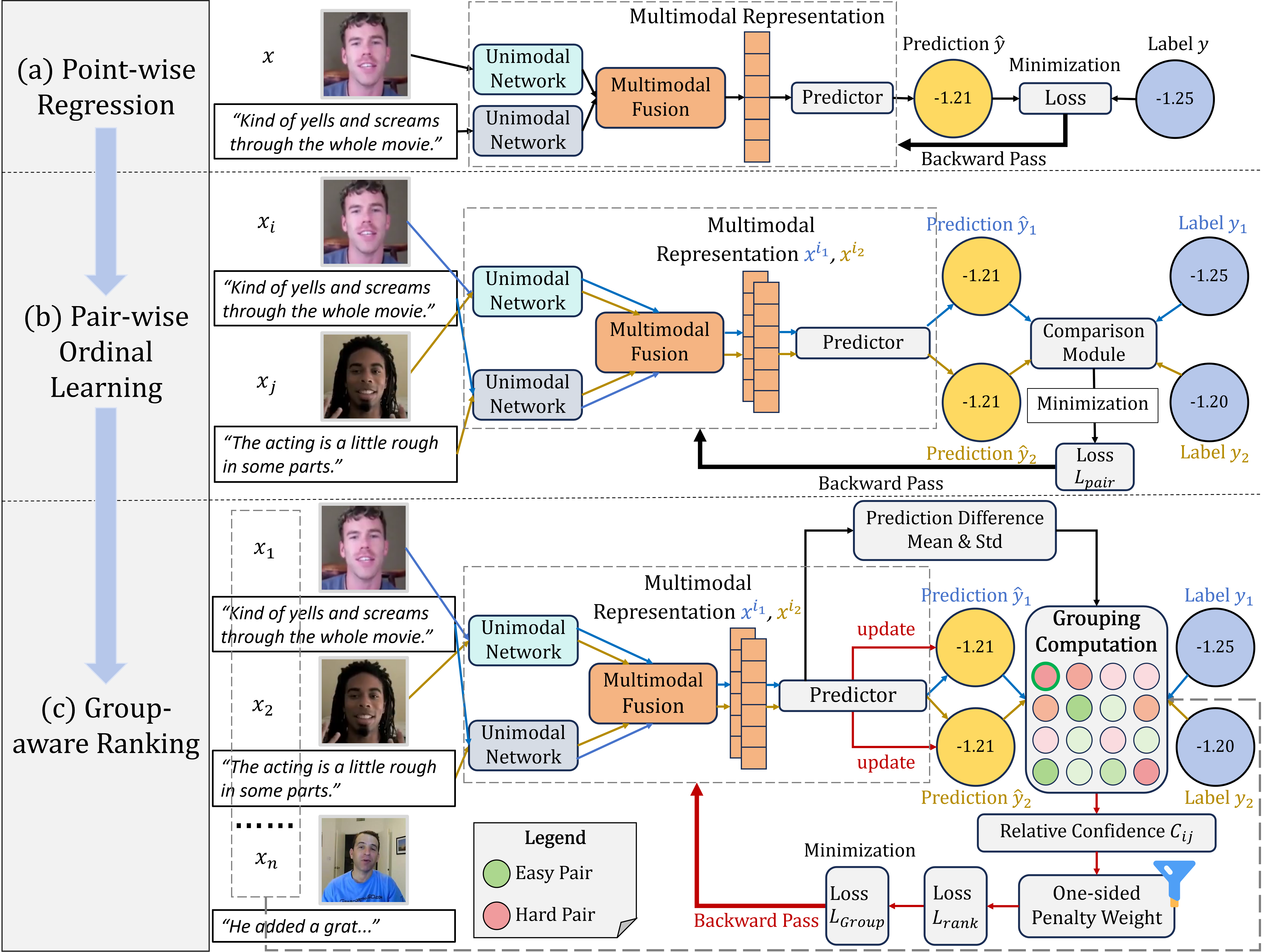}
\caption{A conceptual comparison of different learning paradigms in MSA.}
\label{fig:intro}
\end{figure}

Despite the progress achieved by multimodal transformers~\cite{vaswani2017attention,tsai2019multimodal,cheng2021multimodal} and pre-trained encoders such as BERT~\cite{devlin-etal-2019-bert}, DeBERTa~\cite{He2020DeBERTaDB} and ALBERT~\citep{Lan2019ALBERTAL}, traditional Multimodal Sentiment Analysis (MSA)~\cite{zadeh2016multimodal} frameworks predominantly rely on point-wise regression. As shown in Figure~\ref{fig:intro}, these methods~\cite{hou2019deep,li2023decoupled,Gao_2024_CVPR,Fang_2025_CVPR} compute the loss between continuous labels and predicted scores, treating each instance independently. However, forcefully assigning absolute cardinal values to subjective affective concepts is neither noise-resistant nor aligned with human perception~\cite{yannakakis2018ordinal}. Because such objectives neglect relational ordering, models lack structural constraints to anchor ambiguous samples. Consequently, point-wise models are prone to overfitting subjective label noise rather than capturing the true ordinal hierarchy, often resulting in unstable predictions and arbitrary polarity flips near neutral boundaries.

To overcome the limitations of point-wise regression, recent works have pivoted towards ordinal learning, typically implemented via pair-wise constraints. By evaluating pairs of samples rather than isolated instances, these models are encouraged to preserve the monotonic relative relationships between sentiments (e.g., ensuring sample $i$ is ranked higher than sample $j$). Works like Trustworthy Multimodal Sentiment Ordinal Network (TMSON)~\cite{tmson} and Multimodal Ordinal Affective Computing (MOAC)~\cite{mai2025learningbycomparing} advance MSA by incorporating ordinal reasoning. However, these pair-wise methods introduce two critical limitations: First, they rely on uniform weighting, treating all pair-wise relationships equally. This allows easily distinguishable pairs to dominate the loss, overwhelming the nuanced signals from hard-to-rank (including difficult and confusable) samples. Second, they employ static ordinal constraints (margins), which fail to reflect the varying semantic distances between sentiment groups. For example, the semantic gap between slightly positive (+0.5) and strongly positive (+3.0) is treated indistinctly from the gap between neutral (0) and slightly positive (+0.5). This semantic distance information remains underexploited~\cite{posner2005circumplex}.

Unlike early psychological theories that classify affect into distinct categories~\cite{Ekman01051992}, Russell’s circumplex model~\cite{russell1980circumplex} conceptualizes affective states along a continuous two-dimensional spectrum. Modern cognitive research extends this paradigm, demonstrating that emotional experiences actually inhabit a much higher-dimensional and fuzzy continuous space~\cite{cowen2017self}, inherently resisting sharp categorical boundaries. Rooted in the cognitive foundation of the Law of Comparative Judgment~\cite{thurstone2017law}, humans naturally navigate this complex space by evaluating psychological values through relative comparisons rather than absolute scales. Furthermore, grounded in the Free-Energy Principle~\cite{friston2010free} from neuroscience, the human brain actively allocates more cognitive resources to process ambiguous and highly uncertain affective stimuli. Inspired by how humans naturally process affect ambiguously, gauge affective semantic distances, and allocate more cognitive resources to hard-to-rank samples, we propose a two-stage Group-wise Ranking and Calibration Framework (GRCF), which simultaneously preserves relative ordinal structure and ensures absolute score calibration.

In Stage 1, GRCF establishes a relative ordinal structure via a Group-aware Ranking Loss. First, to address the issue of uniform weighting, we introduce the Adaptive Error-weighting mechanism. By leveraging batch-level prediction statistics, this mechanism functions as a continuous, pair-wise hard-example mining strategy. Rather than simply assessing whether sample $i$ is greater than sample $j$, it evaluates the pair-wise prediction gap against batch statistics, transforming this relative prediction error into a penalty weight. This mechanism forces the optimization process to prioritize hard-to-rank pairs, filtering out gradient noise from easy pairs. Second, to resolve the issue of the static margin, the proposed Dynamic Margin mechanism utilizes overlapping ordinal groups to model semantic ambiguity and homogeneity near decision boundaries by allowing one sample to belong to multiple groups. By computing the margin based on the group memberships, it ensures that the optimization standard scales proportionally with the true semantic gap. 

In Stage 2, a calibration process refines the learned manifold. Building upon the robust structure established in Stage 1, this stage anchors the absolute prediction magnitudes to the annotated sentiment scales. By employing differential learning rates, we ensure GRCF achieves precise value alignment while preserving the previously learned ordinal relationships.

Experiments on regression tasks demonstrate that GRCF achieves state-of-the-art (SOTA) results. GRCF also shows strong generalizability when extended to binary classification tasks. Then, statistical experiments define the ``hardness'' of a pair as difficulty or confusability. Our main contributions are summarized as follows:
\begin{itemize}
    \item We devise the Error-weighting mechanism, which controls the model to focus on hard-to-rank samples.
    \item We adopt a Dynamic Margin realized through overlapping intervals, capturing both the semantic ambiguity and homogeneity near decision boundaries, ensuring that the margin scales proportionally with the true semantic gap.
\end{itemize}

\section{Related Work}
\label{sec:related}
\subsection{Point-wise MSA methods}
Numerous point-wise MSA methods aim to infer human affective states by integrating textual, acoustic, and visual signals.
Early fusion-based methods ~\cite{hou2019deep,poria2017review,sun2022cubemlp,zhu2023multimodal,liang2019learning} such as Tensor Fusion
Network (TFN)~\cite{zadeh2017tensor} and Memory
Fusion Network (MFN)~\cite{zadeh2018memory} explored tensor interactions and memory mechanisms to model temporal dependencies.
Transformer-based models~\cite{rahman2020integrating,li2023interpretable} like Memory
Fusion Network (MulT)~\cite{tsai2019multimodal} later introduced cross-modal attention to enable fine-grained contextual alignment.
With the rise of pretrained language models, recent approaches leverage pre-trained text encoders such as BERT~\cite{devlin-etal-2019-bert}, DeBERTa~\cite{He2020DeBERTaDB} and ALBERT~\citep{Lan2019ALBERTAL}, coupled with lightweight acoustic and visual branches.
Many other models are also devised based on pretrained language model, e.g., Decoupled Multimodal Distillation (DMD)~\cite{li2023decoupled}, Embracing Aleatoric Uncertainty (EAU)\cite{Gao_2024_CVPR} and Modality-Specific Enhanced Dynamic Emotion Experts (EMOE)~\cite{Fang_2025_CVPR}.

Despite progress, existing point-wise models fail to capture the ordinal nature of human sentiment analysis.

\subsection{Ordinal and Pair-wise Ranking Learning}
Sentiment prediction can be naturally formulated as an ordinal regression problem, where preserving sample order provides stronger supervision than discrete classification.
Early pair-wise ranking models such as RankNet~\cite{burges2005learning} and LambdaRank~\cite{6287370} learn monotonic score differences to maintain relative ordering. Xie \textit{et al.}~\cite{tmson} devised TMSON, proposing a novel multimodal fusion framework that combines modality-specific uncertainty modeling with ordinal regression to achieve more reliable sentiment analysis in the ordinal sentiment space.
Recently, Mai \textit{et al.}~\cite{mai2025learningbycomparing} introduced MOAC, which formulates MSA as an ordinal learning problem by performing label-level and feature-level comparisons.
Their approach demonstrates that explicitly modeling ordinal relations improves the interpretability and stability of multimodal sentiment models.

However, these pair-wise methods rely on uniform weighting and static margins, which overlooks how humans naturally gauge varying semantic distances, and actively allocate more cognitive resources to process ambiguous or uncertain affective stimuli.

\section{Methodology}
\label{sec:methodology}
\begin{figure*}[t]
% \vspace{-0.2em}
\centering
\includegraphics[width=\linewidth]{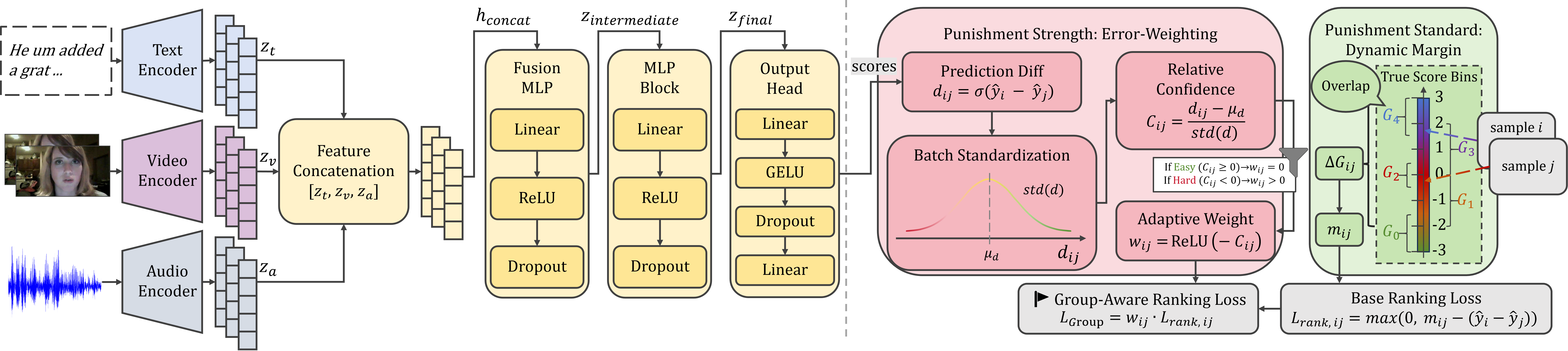}
% \vspace{-0.5em}
\caption{The forward pass of GRCF and core loss calculation which decouples the punishment standard from the strength.}
% \vspace{-0.8em}
\label{fig:framework}
\end{figure*}

\subsection{Multimodal Encoder Architecture}
\label{sec:encoder_arch}
\paragraph{(a) Unimodal Encoders: }Each utterance's text $T$ is processed by a text encoder to produce a global textual representation $\mathbf{z}_t$ (using the [CLS] token output). For the audio $A$ and video $V$ modalities, $\mathbf{z}_a$ and $\mathbf{z}_v$ are produced via attention pooling:

\vspace{-0.4em}
\begin{equation}
\mathbf{z}_a = \sum_{i=1}^{L_a} \alpha_i^{(a)} \mathbf{A}_i, \quad 
\alpha_i^{(a)} = \frac{\exp(\mathbf{w}_a^\top \mathbf{A}_i)}{\sum_{j=1}^{L_a} \exp(\mathbf{w}_a^\top \mathbf{A}_j)},
\label{eq:za_attention}
\end{equation}
and analogously for $\mathbf{z}_v$.

\paragraph{(b) Multimodal Fusion: }The goal is to produce an intermediate representation, $\mathbf{z}_{\text{intermediate}}$, which is then fed into a shared unified encoder. For continuous sentiment, a holistic understanding of the context and intensity from all modalities is paramount. Regression tasks, which aim to produce a single, continuous-like score, inherently require the integration of all modalities into one unified judgment, rather than modeling discrete conflicts between them. Consequently, for this type of holistic integration task, a simple and efficient ``Concatenate-then-MLP'' architecture is effective. First, the unimodal representations are concatenated:

\vspace{-0.4em}
\begin{equation}
\mathbf{h}_{\text{concat}} = [\mathbf{z}_t; \mathbf{z}_a; \mathbf{z}_v].
\end{equation}
This combined vector is then passed through a dedicated fusion MLP to create the intermediate representation:

\vspace{-0.4em}
\begin{equation}
\mathbf{z}_{\text{intermediate}} = \operatorname{MLP}_{\text{fuse}}(\mathbf{h}_{\text{concat}}).
\end{equation}
$\mathbf{z}_{\text{intermediate}}$ captures the salient multimodal information. It is passed through a unified encoder (an MLP block) to create the definitive representation $\mathbf{z}_{\text{final}}$, which is then fed into the regression head. To predict a continuous score $\hat{y}$, the head (a two-layer MLP) constrains the output using a scaling factor $\alpha$ (e.g., 3.5) and a hyperbolic tangent function to match the annotation range (e.g. , [-3, 3]):

\vspace{-0.4em}
\begin{equation}
\hat{y} = \alpha \times \tanh(f_{\operatorname{reg}}(\mathbf{z}_{\text{final}})).
\end{equation}
\subsection{Stage~1: Structural Foundation}
The goal of Stage 1 is to address the problem that traditional regression neglects relational ordering. To this end, we apply a unified loss ($\mathcal{L}_{\text{group}}$) directly at the final output layer. Our hypothesis is that by utilizing Adaptive Error-weighting to implicitly concentrate optimization on hard-to-rank pairs (acting as the ``Punishment Strength''), the model is compelled to learn a robust and well-separated ordinal manifold. This process is guided by the Dynamic Margin, which serves as the ``Punishment Standard''. The Stage 1 objective integrates three key components:

\paragraph{(a) Group-aware Ranking Loss ($\mathcal{L}_{\text{group}}$): }This is the core loss for modeling ordinality. Its design is a multi-step process:

\noindent \textbf{Defining Overlapping Ordinal Groups ($g(\cdot)$): }To enforce the ``semantically proportional'' structure, we model the fuzzy and continuous nature of human sentiment by mapping a continuous score $s_i$ to a \emph{set} of ordinal and overlapping groups $g(s_i)$. 
\label{sec: overlap-group}

Determining the optimal number of groups $K$ involves a trade-off defined by an information efficiency criterion. A larger $K$ increases the information capacity (entropy) to express fine-grained sentiment differences, but if $K$ is too large, the group span $\Delta_K = \frac{S_{max}-S_{min}}{K}$ will drop below the dataset's inherent annotation noise floor ($\epsilon_y$). This implies a theoretical maximum resolvable sentiment limit $K_{max} \approx \frac{S_{max}-S_{min}}{\epsilon_y}$. To ensure robust optimization, $\Delta_K$ must remain comfortably above $\epsilon_y$. Guided by this criterion and using historical SOTA MAE as a proxy for $\epsilon_y$, we can safely define $K=5$ groups using overlapping intervals. 

This design allows a single sample to belong to multiple groups, which serves two key purposes: it captures the inherent ambiguity of a single score on a boundary (for instance, a score of $s_i = 0.2$ is assigned to both $G_2$ and $G_3$), while also ensuring that numerically-close boundary samples (e.g., -1.8 and -1.6) share same group membership, correctly identifying them as semantically similar. Specifically, with scores ranging from -3 to 3, the groups are defined as: $G_0$ [-3.0, -1.5], $G_1$ [-2.0, 0.0], $G_2$ [-0.5, 0.5], $G_3$ [0.0, 2.0], and $G_4$ [1.5, 3.0].

\noindent \textbf{Calculating the Dynamic Margin: }For a pair-wise samples $(i, j)$ with scores $s_i > s_j$, the group difference $\Delta G_{ij}$ is defined as the maximum semantic distance between their groups:

\vspace{-0.4em}
\begin{equation}
\label{eq:G}
\Delta G_{ij} =
\begin{cases}
0, & \text{if } G(s_i) \cap G(s_j) \neq \emptyset \\
\max\limits_{G_a \in G(s_i), G_b \in G(s_j)} |G_a - G_b|, & \text{otherwise}.
\end{cases}
\end{equation}
Consequently, a comparison between an ambiguous sample (e.g., $s_i=0.2$ in $G_2, G_3$) and a stable sample (e.g., $s_j=2.5$ in $G_4$) is robustly constrained by the largest possible semantic gap (e.g., $|4-2|=2$). If two samples share any group (e.g., $s_i=-1.8$ and $s_j=-1.6$, which are both in $G_0$ and $G_1$), their $\Delta G_{ij}$ becomes 0. The Dynamic Margin $m_{ij}$ is then defined based on this group difference $\Delta G_{ij}$:

\vspace{-0.4em}
\begin{equation}
m_{ij} =
\begin{cases}
m_{\text{intra}}, & \Delta G_{ij}=0,\\[2pt]
m_{\text{base}}+\Delta G_{ij}m_{\text{step}}, & \text{otherwise},
\end{cases}
\end{equation}
where $m_{\text{intra}}$ acts as a small base margin for samples within the same group to maintain a minimal local separation, $m_{\text{base}}$ is the foundational margin for inter-group pairs, and $m_{\text{step}}$ is the incremental margin multiplier scaled by the semantic distance $\Delta G_{ij}$. This mechanism is the solution to the ``static constraint'' problem. Instead of treating all pairs uniformly, our margin now scales proportionally with the true semantic gap defined by the groups. This teaches GRCF that inter-group pairs with larger semantic gaps must have a larger predicted score difference than intra-group pairs with small gaps. The base ranking loss for the pair $(i,j)$ is then:

\vspace{-0.4em}
\begin{equation}
\mathcal{L}_{\text{rank},ij}=\max\!\bigl(0,\,m_{ij}-(\hat{y}_i-\hat{y}_j)\bigr).
\end{equation}

\noindent \textbf{Applying Adaptive Error-weighting: }
Having established a standard for separation ($m_{ij}$), we now introduce an adaptive strength for optimization ($w_{ij}$) using a dynamic re-weighting principle. First, we compute the raw prediction difference $d_{ij}$ for a pair, representing the model's absolute confidence in their relative order:

\vspace{-0.4em}
\begin{equation}
    d_{ij} = \sigma(\hat{y}_i - \hat{y}_j).
\end{equation}
Instead of applying a symmetric loss to all pairs, we compare this difference against the global batch distribution to identify which pairs the model is struggling with. We calculate a relative confidence score $C_{ij}$ by standardizing $d_{ij}$ within the current batch:

\vspace{-0.4em}
\begin{equation}
    C_{ij} = \frac{d_{ij} - \mu_d}{\operatorname{std}(d)},
\end{equation}
where $\mu_d$ and $\operatorname{std}(d)$ are the mean and standard deviation of $d_{ij}$ across all valid pairs in the batch. 

A positive $C_{ij}$ indicates an ``easy pair'' where the model's confidence exceeds the batch average, while a negative $C_{ij}$ signifies a ``hard-to-rank pair''. To filter out gradient noise from easy pairs, we transform this relative confidence metric into an adaptive, one-sided penalty weight $w_{ij}$:

\vspace{-0.4em}
\begin{equation}
    w_{ij} = \operatorname{ReLU}(-C_{ij}).
\end{equation}
The Error-weighting mechanism with a one-sided penalty acts as an adaptive filter that mimics this biological process by allocating higher optimization penalties to hard-to-rank pairs. It effectively up-weights pairs with negative relative confidence to force the optimization process to focus exclusively on hard-to-rank pairs.

\vspace{-0.4em}
\begin{equation}
\mathcal{L}_{\text{group}}=\frac{1}{N_p}\sum_{s_i>s_j}w_{ij} \cdot \mathcal{L}_{\text{rank},ij}.
\label{eq:grpo_loss}
\end{equation}
This allows GRCF to learn a smoother and more semantically coherent ordinal manifold.

A critical consideration is how this Adaptive Error-weighting specifically benefits multimodal sentiment analysis, as opposed to being a generic unimodal optimization trick. In multimodal scenarios, the most challenging samples, which generate the highest prediction errors and thus receive the highest adaptive weights ($w_{ij}$), are predominantly instances characterized by cross-modal conflict or subtle, asynchronous non-verbal cues, such as a speaker smiling while expressing negative semantics.

Unlike standard point-wise regression that passively averages out these conflicting multimodal signals, our error-weighted Dynamic Margin acts as an implicit cross-modal conflict resolver. By actively forcing the optimization process to focus on these hard-to-rank pairs, the unified encoder is compelled to learn nuanced cross-modal interactions rather than relying on a dominant modality like text. Consequently, while the mathematical formulation of GRCF is modality-agnostic, its dynamic hard-mining behavior is intrinsically tailored to untangle the complex, contradictory affective signals unique to multimodal data.

% \vspace{-0.5cm}
\paragraph{(b) Distribution Regularization Loss ($\mathcal{L}_{\text{reg}}$): }To ensure the learned relative rankings do not ``drift'' in absolute terms, we regularize the predicted batch distribution ($\mathbf{\hat{y}}$) to match the ground-truth batch distribution ($\mathbf{s}$). This loss penalizes deviations in mean ($M$) and standard deviation ($D$) beyond a margin ($\gamma$):

\vspace{-0.4em}
\begin{equation}
\mathcal{L}_{\text{reg}} = \left[(M_p - M_r)^2 - \gamma\right]_+ + \left[(D_p - D_r)^2 - \gamma\right]_+.
\end{equation}

\paragraph{(c) Boundary Loss ($\mathcal{L}_{\text{bound}}$): }Finally, to ensure predictions remain within the valid annotation range $[-S, S]$, we add a boundary loss:

\vspace{-0.4em}
\begin{equation}
\mathcal{L}_{\text{bound}} = \frac{1}{N} \sum_{i=1}^{N} \max(0, |\hat{y}_i| - S).
\label{eq:boundary_loss}
\end{equation}
The total loss for Stage 1 is a weighted sum:

\vspace{-0.4em}
\begin{equation}
\mathcal{L}_{S1} = \lambda_1 \mathcal{L}_{\text{group}} + \lambda_2 \mathcal{L}_{\text{reg}} + \lambda_3 \mathcal{L}_{\text{bound}}.
\end{equation}

\subsection{Stage~2: Fine-Tuning}
While Stage 1 establishes a robustly ordinal space, it inherently suffers from a ``score offset'' problem. Instead of employing a rigid point-wise constraint (e.g., Mean Absolute Error, MAE), Stage 1 relies exclusively on relative ordering ($\mathcal{L}_{\text{group}}$) and soft macroscopic alignment ($\mathcal{L}_{\text{reg}}$). Consequently, while the model successfully learns the correct semantic distances between samples, it lacks a definitive anchor for absolute score values (e.g., the exact neutral $0$ point). 

Although the scaled $\operatorname{tanh}$ function and Boundary Loss ($\mathcal{L}_{\text{bound}}$) prevent predictions from exceeding the valid annotation range, the entire learned ordinal manifold remains free to ``drift'' within these bounds. Therefore, Stage 2 primarily corrects this offset by introducing direct MAE-driven calibration, anchoring the learned relative manifold to exact absolute scores. To prevent catastrophic forgetting of the learned structure, we employ differential learning rates, applying a higher rate to the fusion/output layers and a lower rate to the foundational encoders.
To preserve the learned structure, $\mathcal{L}_{\text{mae}}$ is integrated with the $\mathcal{L}_{\text{group}}$ and $\mathcal{L}_{\text{bound}}$ losses from Stage 1, which now maintain the ranking ability and act as regularizer:

\vspace{-0.4em}
\begin{equation}
\mathcal{L}_{S2}= \beta_1 \mathcal{L}_{\text{mae}} + \beta_2 \mathcal{L}_{\text{group}} + \beta_3 \mathcal{L}_{\text{bound}}.
\end{equation}
The weight for MAE ($\beta_1$) is set higher, providing a strong gradient for calibration while others preserve the ordinal structure.

\subsection{Extending to Binary Classification Tasks}
GRCF is inherently suitable for regression because continuous labels provide a naturally dense space for evaluating relative prediction differences. However, adapting GRCF to binary classification presents distinct challenges. First, the task relies on sparse, discrete labels (e.g., Non-sarcastic/Sarcastic), which limits the fine-grained granularity required by our Adaptive Error-weighting mechanism. Second, in binary classification, the existence of only two mutually exclusive classes invalidates the Dynamic Margin mechanism, which relies on multiple fine-grained, overlapping ordinal groups (e.g., $G_0$ to $G_4$) to calculate semantic distance.

Therefore, a principled adaptation was required. Instead of applying an Ordinal Loss with an overlapping-group-based Dynamic Margin, we decomposed the group-wise philosophy into a Separation Loss to push inter-class clusters apart and a Compactness Loss to pull intra-class samples together.

\section{Experiments}
\label{sec:experiments}
\subsection{Implementation Details}
We use \texttt{DeBERTa-v3-base} as the default text backbone; acoustic and visual features are aggregated via attention pooling and projected to a shared latent space before fusion. Training uses the AdamW optimizer with mixed precision, cosine learning-rate decay, gradient clipping, and distributed data parallelism. We conduct a two-part hyperparameter search via Optuna~\cite{akiba2019optunanextgenerationhyperparameteroptimization}. Stage 1 maximizes Pair-wise Accuracy (PairwiseAcc), while Stage 2 minimizes MAE. We defined a reasonable range and applied a random search with 50 trials. We trained the model on a server with an NVIDIA RTX 4090 GPU and 377 GB of system RAM.

\begin{figure}[t]
\centering
\includegraphics[width=\linewidth]{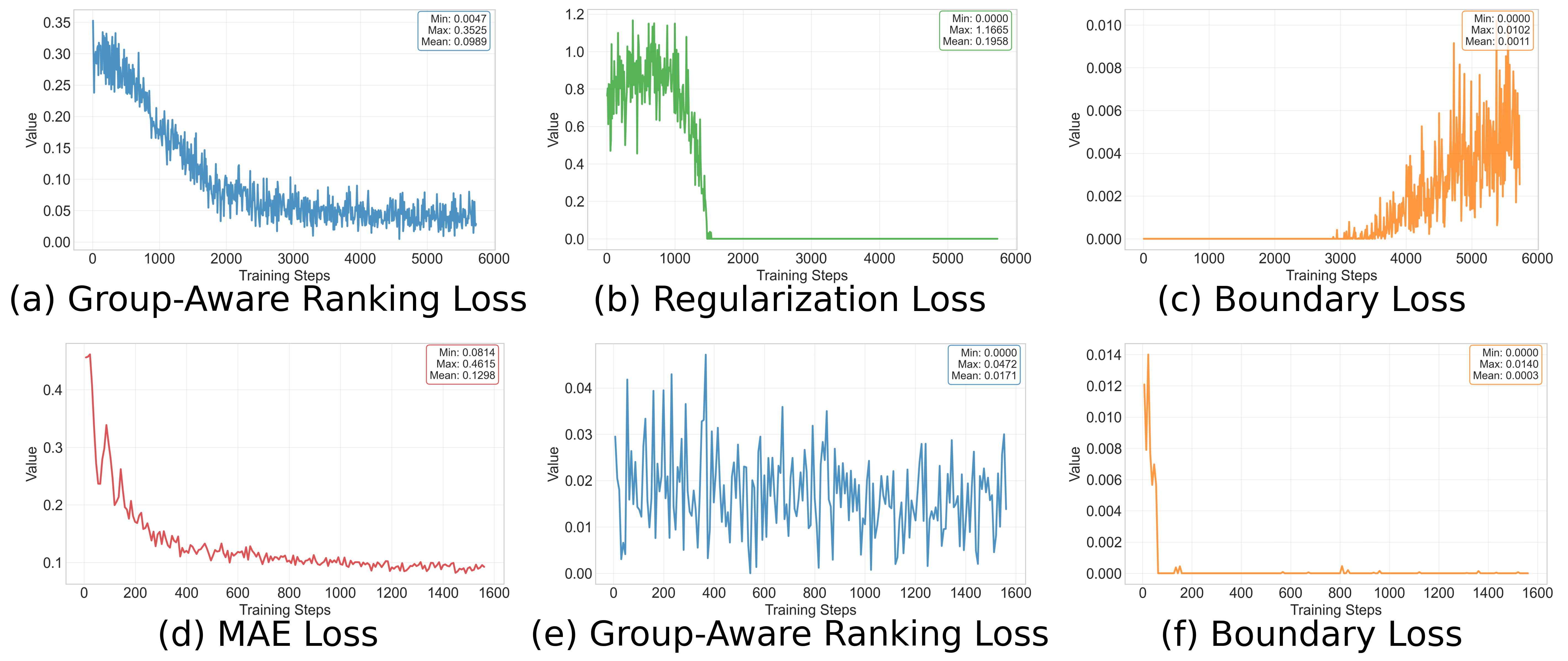}
\caption{Training loss curves on CMU-MOSI dataset. (Top) Stage 1 loss. (Bottom) Stage 2 loss.}
\label{fig:loss}
\end{figure}

As shown in Figure~\ref{fig:loss}, in Stage 1 training, the Group-aware Ranking Loss decreases steadily, and the Distribution Regularization Loss is active early in the process. Moreover, the Boundary Loss only activates during the later stages, which is a direct consequence of ordinal learning. As the ranking objective continuously pushes the pairs apart to establish clear relative distances, the overall score distribution naturally expands outward. Once the predictions hit the boundaries, the Boundary Loss actively triggers to penalize and bound these values. In Stage 2 training, the primary MAE loss for calibration trends downward, the Group-aware Ranking Loss remains stable, preserving the learned ranking, and the Boundary Loss is minimal, indicating few out-of-range predictions. 

Notably, in Figure~\ref{fig:loss} (d), the training MAE  converging to a minimum, which appears to contrast to the test MAE (0.581). Actually, this indicates that GRCF enters a benign interpolation regime. Because the optimal group structure established in Stage 1 robustly constrains the global ordinal manifold, Stage 2 interpolates these absolute scores locally via high-frequency adjustments without disrupting the global topology. Consequently, this prevents detrimental overfitting to intrinsic label noise and safely maps test samples to a robust semantic manifold.

\subsection{Main Results on Sentiment Analysis}
Table \ref{tbase_1} presents the performance of GRCF on CMU-MOSI~\cite{zadeh2016multimodal} and CMU-MOSEI~\cite{zadeh2018multimodal} datasets. On the CMU-MOSI dataset, GRCF achieves SOTA results on all evaluation metrics. Conversely, on the CMU-MOSEI dataset, GRCF outperforms existing methods across all metrics, reaching SOTA levels consistently. 

Table~\ref{tab:SIMSResult} shows the results on the Chinese-specific CH-SIMS v2~\cite{liu2022make} dataset. To ensure a rigorous evaluation regarding backbone dependency and fair comparisons with prior works, we evaluate GRCF variants across different text encoders. While default \texttt{DeBERTa-v3} already achieves highly competitive results, utilizing language-aligned encoders like \texttt{mDeBERTa-v3} and \texttt{BERT-chinese} yields massive performance leaps. GRCF (\texttt{BERT-chinese}) establishes a completely new SOTA across all metrics, substantially outperforming the previous baselines (e.g., Acc5 58.4\%, and Corr 0.744). This finding strongly proves two points: First, GRCF is robust and encoder-agnostic, capable of achieving peak performance under the BERT backbone setting used by earlier SOTA methods. Second, it demonstrates that GRCF inherently amplifies the representation power of the chosen encoder, rather than relying exclusively on DeBERTa's capabilities.
\begin{table*}[t]
\centering
\small
 \caption{ \label{tbase_1}The comparison with baselines on CMU-MOSI and CMU-MOSEI. Acc2 and F1-score are calculated exclusively on positive and negative samples, with zero-valued instances excluded from the evaluation. The results labeled with $^{\dag}$ are obtained from original papers, and others are obtained from our experiments. Best results are in bold, and second-best are underlined.
 }
 \begin{tabular}{l|c|c|c|c|c|c|c|c|c|c}
  \noalign{\hrule height 1pt} 
       \multirow{2}{*}{\textbf{Baseline Model}} & \multicolumn{5}{c|}{\textbf{CMU-MOSI}} & \multicolumn{5}{c}{\textbf{CMU-MOSEI}}  \\
 \cline{2-11}
& Acc7$\uparrow$  & Acc2$\uparrow$ & F1-score$\uparrow$ & MAE$\downarrow$ & Corr$\uparrow$ & Acc7$\uparrow$  & Acc2$\uparrow$ & F1-score$\uparrow$ & MAE$\downarrow$ & Corr$\uparrow$\\
\hline
\textit{Point-wise} \\
\hline
    Self-MM \citep{mmsa} &  45.8 &   84.9  &   84.8  & 0.731  & 0.785  &  53.0 &   85.2  &   85.2  &  0.540 & 0.763 \\
AtCAF$^{\dag}$ \citep{HUANG2025102725}  & 46.5  & 88.6  &  88.5  &  0.650 & 0.831  & \underline{55.9}  & 87.0  &  86.8 & \underline{0.508} & 0.785 \\
DLF$^{\dag}$ \citep{wang2025dlf}  & 47.1  & 85.1  &  85.0  &  0.731 & 0.781  & 53.9  & 85.4  &  85.3 & 0.536 & 0.764 \\
KuDA \cite{feng-etal-2024-knowledge} & 47.1 & 86.4 & 86.5 & 0.705 & 0.795 & 52.9 & 86.5 & 86.6 & 0.529 & 0.776 \\
DEVA$^{\dag}$ \citep{deva} & 46.3  & 86.3 & 86.3  &  0.730 & 0.787  & 52.3  &  86.1 & 86.2 & 0.541  & 0.769 \\
C-MIB \citep{MIB} &  47.7 &   87.8 &  87.8  & 0.662  & 0.835  & 52.7 &  86.9  & 86.8 & 0.542  &  0.784\\
ITHP \citep{ithp} &  47.7 &   88.5 &   88.5  & 0.663  & 0.856  & 52.2  &   87.1  & 87.1 &  0.550 & 0.792 \\
Multimodal Boosting \citep{10224356} & \underline{49.1}  &  88.5 & 88.4  &  0.634 & 0.855  & 54.0  &  86.5 & 86.5 & 0.523  & 0.779 \\
CaMIB$^{\dag}$ \citep{DBLP:journals/corr/abs-2509-21805} & 48.0  &  \underline{89.8} & \underline{89.8}  &  0.616 & \underline{0.857}  & 53.5  &  87.3 & 87.2 & 0.517 & 0.788 \\
DMD \citep{li2023decoupled} &44.9	&84.3	&84.3	&0.726 	&0.788  & 
 52.8	&84.6	&84.6	&0.538&0.768\\
EMOE \citep{Fang_2025_CVPR} &45.2	&84.8	&84.8	&0.723 	&0.790   &  52.5	&85.0	&85.0	&0.542 	&0.760  \\
TSD$^{\dag}$ \citep{meng2026trisubspacesdisentanglementmultimodalsentiment} & 49.0 & 86.5 & 86.6 & 0.691 & - & 54.6 & 86.2 & 86.2 & 0.525 & - \\
DecAlign$^{\dag}$ \citep{qian2025decalignhierarchicalcrossmodalalignment} & - & 85.8 & 85.8 & 0.735 & - & - & 86.5 & 86.1 & 0.543 & - \\
\hline
\textit{Pair-wise} \\
\hline
TMSON$^{\dag}$ \citep{tmson} & 47.4  &  87.2 & 87.2  &  0.687 & 0.809  & 55.6  &  86.4 & 86.2 & 0.526  & 0.766 \\
MOAC \citep{mai2025learningbycomparing} & 48.6  &  89.0 & 89.0  &  \underline{0.605} & \underline{0.857}  & 54.3  &  87.6 & \underline{87.6} & 0.512  & 0.793 \\
MOL \citep{11493959} & 48.3 & 89.5 & 89.4 & 0.619 & 0.854 & 54.7 & \underline{87.7} & \underline{87.6} & 0.512 & \underline{0.794} \\
% \hline
GRCF (Single-stage Bi-objective) & 45.3 & 88.1 & 88.1 & 0.650 & 0.849 & 56.0 & 86.6 & 86.5 & 0.516 & 0.792 \\
GRCF (Two-stage) & \textbf{49.3}  &   \textbf{90.3} &  \textbf{90.2}  &  \textbf{0.581} & \textbf{0.866}  & \textbf{58.7} &  \textbf{89.1} &   \textbf{89.1} & \textbf{0.461}  &  \textbf{0.835} \\
\noalign{\hrule height 1pt} 
\end{tabular}
\end{table*}

\begin{table}[t]
\centering
\small
\setlength{\tabcolsep}{3pt}
    \caption{The comparison with baselines on CH-SIMS v2.}
    \label{tab:SIMSResult}
        \begin{tabular}{l|c|c|c|c|c|c}
        \noalign{\hrule height 1pt} 
        \multirow{2}{*}{\textbf{Baseline Model}} & \multicolumn{5}{c}{\textbf{CH-SIMS v2}} \\ \cline{2-7}
         & Acc5$\uparrow$ & Acc3$\uparrow$ & Acc2$\uparrow$ & F1-score$\uparrow$ & MAE$\downarrow$ & Corr$\uparrow$  \\ \hline
        % EF-LSTM \cite{williams2018recognizing} & 53.7 & 73.5 & 80.1 & 80.0 & 0.309 & 0.700  \\ 
        % LF-DNN \cite{williams2018dnn} & 51.8 & 71.2 & 77.8 & 77.9 & 0.322 & 0.668 \\
        TFN \cite{zadeh2017tensor} & 53.3 & 70.9 & 78.1 & 78.1 & 0.322 & 0.662 \\
        LMF \cite{liu2018efficient} & 51.6 & 70.0 & 77.8 & 77.8 & 0.327 & 0.651 \\
        MFN \cite{zadeh2018memory} & 55.4 & 72.7 & 79.4 & 79.4 & 0.301 & 0.712 \\
        Graph-MFN \cite{zadeh2018multimodal} & 48.9 & 68.6 & 76.6 & 76.6 & 0.334 & 0.644 \\
        MISA \cite{hazarika2020misamodalityinvariantspecificrepresentations} & 47.5 & 68.9 & 78.2 & 78.3 & 0.342 & 0.671 \\
        MAG-BERT \cite{rahman2020integrating} & 49.2 & 70.6 & 77.1 & 77.1 & 0.346 & 0.641 \\
        Self-MM \cite{mmsa} & 53.5 & 72.7 & 78.7 & 78.6 & 0.315 & 0.691 \\
        MMIM \cite{han-etal-2021-improving} & 50.5 & 70.4 & 77.8 & 77.8 & 0.339 & 0.641 \\
        AV-MC \cite{liu2022make} & 52.1 & 73.2 & 80.6 & 80.7 & 0.301 & 0.721 \\
        KuDA \cite{feng-etal-2024-knowledge} & 53.1 & 74.3 & 80.2 & 80.1 & 0.289 & 0.741 \\
        KAN-MCP\cite{luo2025towards} & \underline{57.3} & \underline{75.0} & 81.6 & 81.7 & \underline{0.281} & \underline{0.742}  \\
        \hline
        GRCF (\texttt{DeBERTa-v3}) & 54.4 & 74.3 & 81.7 & 81.6 & 0.305 & 0.691  \\  
        GRCF (\texttt{mDeBERTa-v3}) & 54.7 & \underline{75.0} & \underline{82.5} & \underline{82.5} & 0.294 & 0.705  \\ 
        GRCF (\texttt{BERT-chinese}) & \textbf{58.4} & \textbf{75.9} & \textbf{83.5} & \textbf{83.5} & \textbf{0.271} & \textbf{0.744}  \\  
        \noalign{\hrule height 1pt}
        \end{tabular}
\end{table}

\subsection{Results on Humor and Sarcasm Detection}
\begin{figure}[t]
\centering
\begin{subfigure}[b]{0.495\linewidth}
    \includegraphics[width=\linewidth]{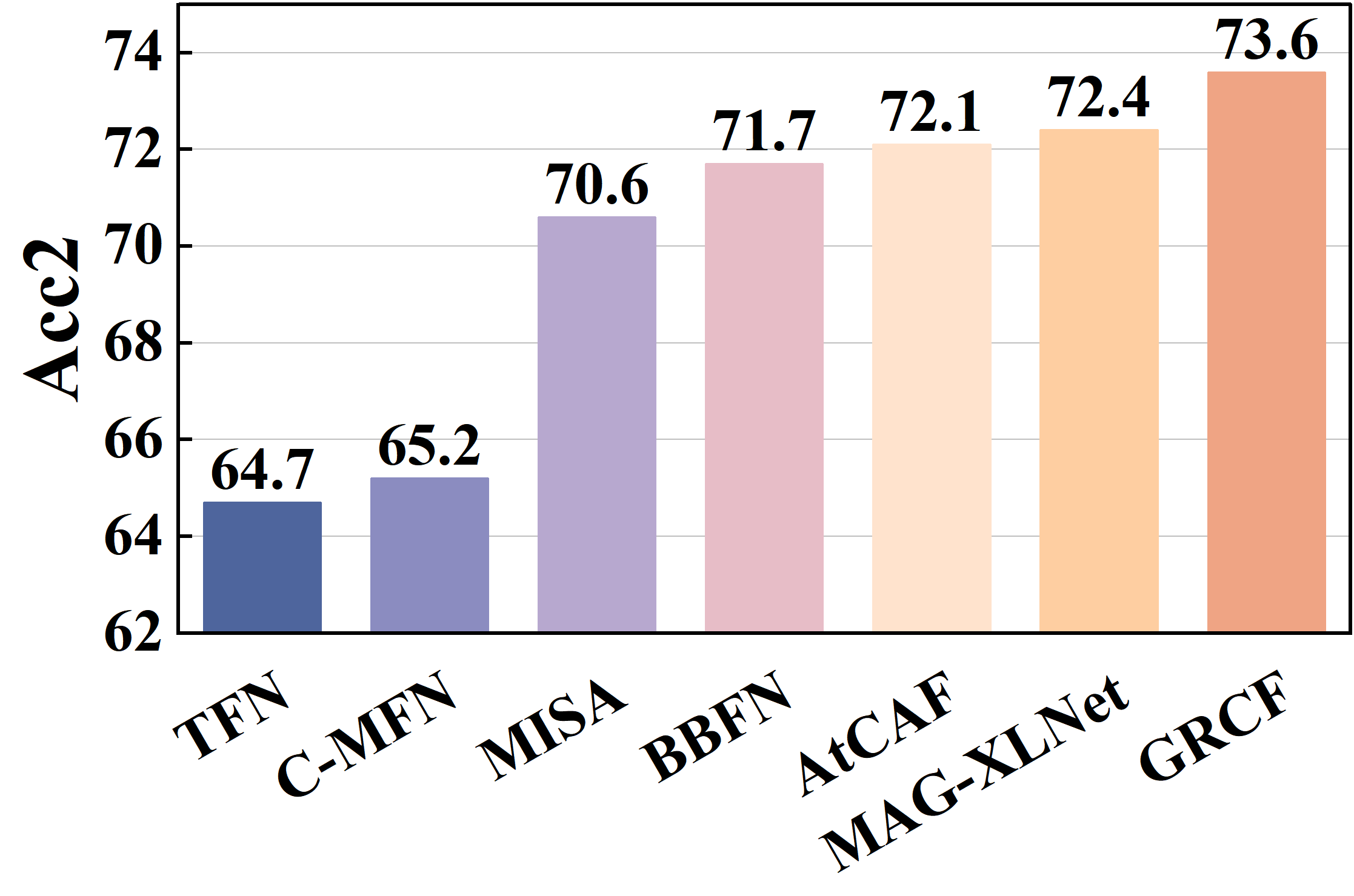}
    \caption{URFUNNY v2}
    \label{fig:mhd}
\end{subfigure}
\hfill % 
\begin{subfigure}[b]{0.495\linewidth}
    \includegraphics[width=\linewidth]{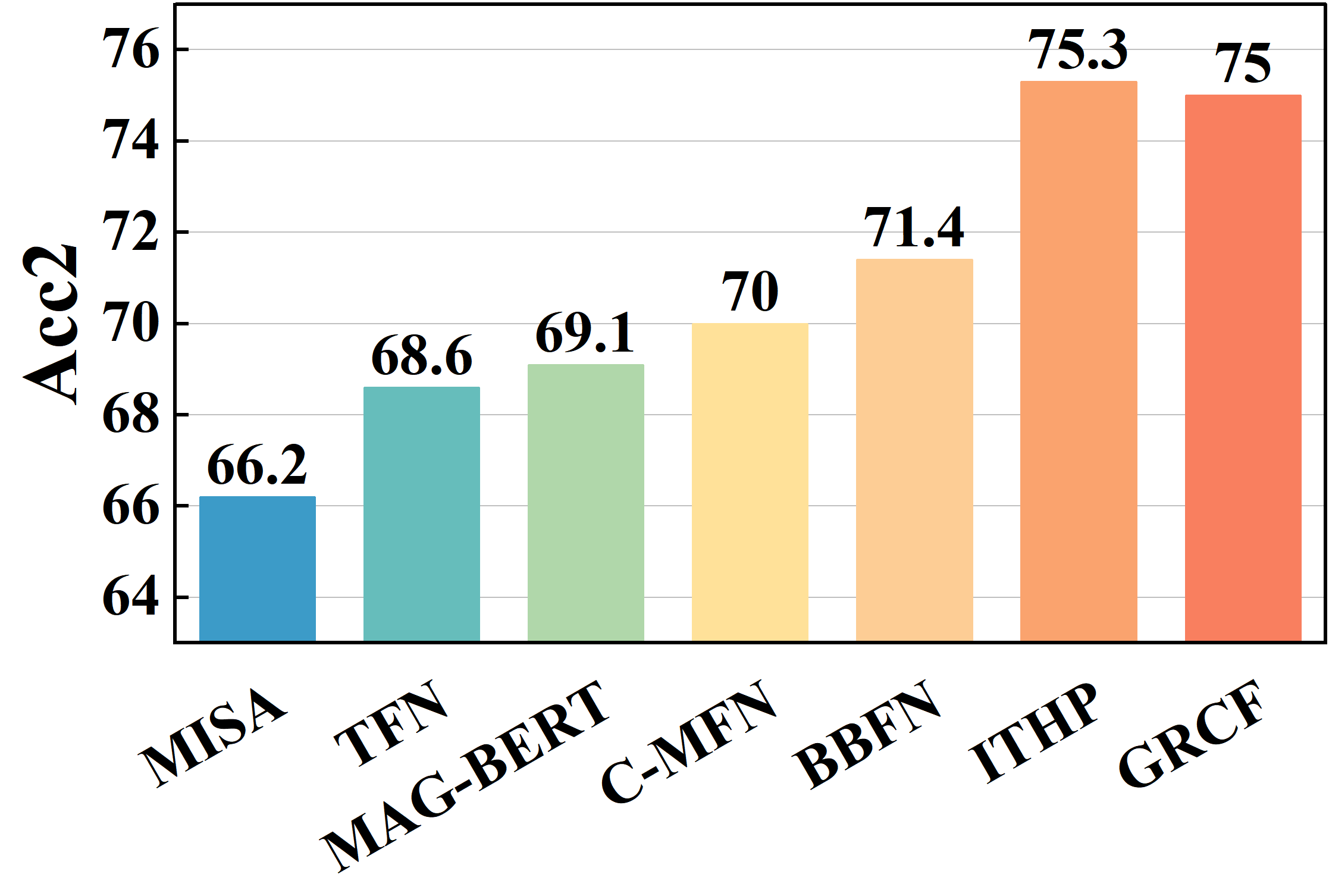}
    \caption{MUStARD}
    \label{fig:msd}
\end{subfigure}
\caption{Results (Acc2) on the classification tasks.}
\label{fig:classification_results}
\end{figure}

We adapt it for the binary classification tasks of Multimodal Humor Detection (MHD) on UR-FUNNY v2~\cite{hasan-etal-2019-ur} and Multimodal Sarcasm Detection (MSD) on MUStARD~\cite{castro-etal-2019-towards}. It achieves an Acc2 of 75.00\% on MUStARD and 73.64\% on UR-FUNNY v2, demonstrating that GRCF generalizes effectively for classification tasks, not just for regression. Results are presented in Figure~\ref{fig:classification_results}.

\subsection{Ablation Experiment}
\begin{table}[t]
\centering
\small
\caption{Ablation study results on different datasets. W/o denotes removing the component.}
\label{tab:ablation_study} 
\begin{tabular}{l|c|c|c|c} 
\noalign{\hrule height 1pt} 
\multirow{2}{*}{\textbf{Method}}  & \multicolumn{2}{c|}{\textbf{CMU-MOSI}} & \multicolumn{2}{c}{\textbf{CMU-MOSEI}} \\
\cline{2-3} \cline{4-5} 
& Acc7$\uparrow$ & Corr$\uparrow$ & Acc7$\uparrow$ & Corr$\uparrow$ \\
\hline 
w/o Error-weighting & 
45.0 & 
0.835 & 
53.9 & 
0.736 
\\

w/o Dynamic Margin & 
46.7 & 
0.849 & 
53.9 & 
0.749 
\\

w/o Stage 1 & 
43.6 & 
0.852 & 
52.9 & 
0.729 
\\

w/o Stage 2 & 
37.7 & 
0.855 & 
42.5 & 
0.695  
\\
\hline 
Full Model & 49.3 & 0.866 & 58.7 & 0.835 
\\
\noalign{\hrule height 1pt} 
\end{tabular}

\end{table}
To validate the necessity of Stage 1 components and the two-stage framework, we conducted an ablation study on CMU-MOSI, with results presented in Table~\ref{tab:ablation_study}.

\paragraph{(a) Stage 1 Components: }Adaptive Error-weighting and the Dynamic Margin are highly effective for datasets with granular sentiment labels like CMU-MOSI and CMU-MOSEI (range $[-3, 3]$). On CMU-MOSI, removing either Error-weighting or the Dynamic Margin causes a substantial performance drop (e.g., -8.72\% and -5.27\% in Acc7, respectively), confirming their importance in establishing a well-separated ordinal manifold. 

\paragraph{(b) Two-stage Framework: }Removing Stage 1 and relying on direct regression causes a drop (e.g., -11.55\% in Acc7 on CMU-MOSI), demonstrating the necessity of the ranking structural foundation. Moreover, removing Stage 2 results in a catastrophic performance collapse, proving that Stage 2 is indispensable for mapping ordinal space to absolute scores. 

To further justify the necessity of the decoupled two-stage design, we compare our framework against a Single-Stage Bi-objective variant, which optimizes both $\mathcal{L}_{\text{group}}$ and $\mathcal{L}_{\text{mae}}$ concurrently (supplemented by $\mathcal{L}_{\text{bound}}$ and $\mathcal{L}_{\text{reg}}$). As demonstrated by the GRCF (Single-Stage) results incorporated in Table~\ref{tbase_1}, joint optimization yields significantly inferior performance across all metrics compared to our proposed two-stage approach. We attribute this to gradient conflicts: the relative ranking objective (which aggressively pushes boundary samples apart) and the absolute calibration objective (which anchors samples to specific numerical values) interfere with each other when optimized simultaneously. 

Furthermore, decoupling improves computational efficiency. Table~\ref{tab:comp_cost} shows that the two-stage model converges rapidly in merely 22 epochs (1816s). In contrast, the single-stage variant struggles to balance the conflicting objectives, requiring over 30 epochs (2971s) to reach a sub-optimal convergence. This proves that the two-stage paradigm is necessary to resolve optimization conflicts.

\begin{figure*}[t]
\centering
\includegraphics[width=\linewidth]{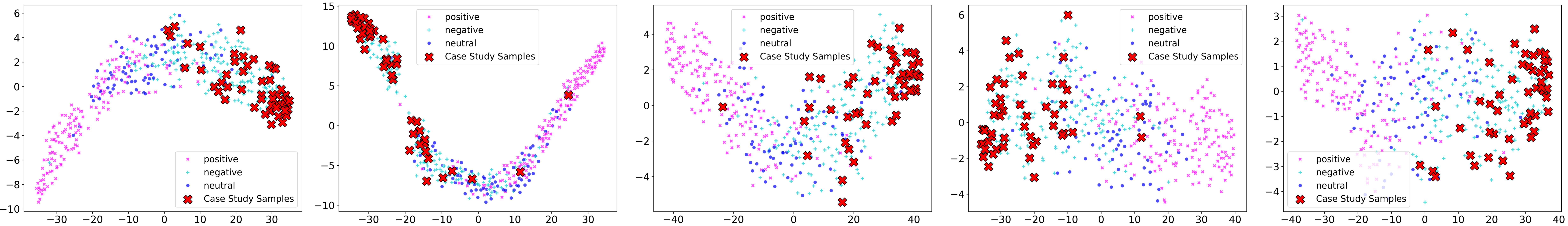}
\caption{T-SNE visualization of ablations on CMU-MOSI. (a) Full model. (b) w/o Error-weighting. (c) w/o Dynamic Margin. (d) w/o Stage 1. (e) w/o Stage 2. }
\label{fig:abl}
\end{figure*}

\subsection{Qualitative Analysis}
To provide an intuitive understanding of how proposed components contribute to predictions, we visualize the internal representation space via t-SNE for full model and four ablations in Figure~\ref{fig:abl}. Instead of examining individual points, we highlight the case study samples whose ground-truth scores are smaller than $-2.0$ as red ``X''s.

\paragraph{(a) Full Model: }It enhances the structure established by Stage 1. The case study cluster in (a) is visibly more compact than the ``ranking-only'' cluster in (e). This shows that $\mathcal{L}_{\text{mae}}$ not only aligns the absolute scores but also acts as a fine-tuning mechanism, further improving the intra-class compactness of the learned manifold.

\paragraph{(b) w/o Error-weighting \& (c) w/o Dynamic Margin: }These reveal why $\mathcal{L}_{\text{group}}$ is effective. Without Error-weighting (b), the optimization is dominated by easy pairs while hard boundary samples are neglected, causing the continuous ordinal spectrum to fracture into disconnected extremes. Without the Dynamic Margin (c), the red ``X'' cluster scatters and ``leaks'' heavily into adjacent zones. This proves Error-weighting ensures a smooth ranking transition by resolving ambiguous middle ground, while the Dynamic Margin enforces strict proportional boundaries.

\paragraph{(d) w/o Stage 1: }It represents a complete structural failure. The case study samples are chaotically scattered across the entire representation space, indiscriminately mixing with ``neutral'' and even ``positive'' samples, demonstrating that $\mathcal{L}_{\text{mae}}$ is incapable of learning a semantically meaningful or ordinally-aware manifold.

\paragraph{(e) w/o Stage 2: }Stage 1 successfully establishes a clear ordinal gradient (placing positive samples on the left and negative on the right). However, the case study cluster remains loosely dispersed, proving Stage 1 builds the foundational ranking structure but requires Stage 2 for tight calibration.

In summary, our components are symbiotic. Stage 1 builds the foundational ordinal structure that direct regression completely misses, while Stage 2 refines this structure, leading to highly cohesive and strictly discriminative representations in the full model.

\subsection{Analysis of Error-weighting Mechanism}
To understand how the Adaptive Error-weighting mechanism focuses on ``hard-to-rank pairs'', we analyze the highest penalty weight samples from CMU-MOSI and MUStARD. We hypothesize that the definition of ``hardness'' is inherently task-dependent:
\begin{itemize}
    \item For MSA, a modality congruence task, the challenge lies in discerning degrees of similarity (e.g., score proximity).
    \item For MSD, a literal meaning subversion task, the challenge is to overcome strong unimodal biases (specifically, the literal meaning of text) via cross-modal contrast.
\end{itemize}

\paragraph{(a) On MSA: Hardness is Driven by Score Proximity:} We conducted an Ordinary Least Squares (OLS) multiple regression analysis to distinguish whether the penalty weight ($w_{ij}$) on CMU-MOSI was driven by score proximity or multimodal modality conflict. 

As shown in Table~\ref{tab:weight_regression_mosi}, when both factors were used as predictors, score proximity emerged as the only statistically significant factor ($p=0.019$). Conflict count was not significant ($p=0.361$), indicating that it provides no independent explanatory power. Thus, in a congruent task like MSA, the true challenge is distinguishing score proximity, and the Error-weighting successfully identifies these semantically close pairs as hard-to-rank cases.

\begin{table}[t]
\small
\centering
\caption{OLS regression analyzing predictors of the penalty weights on CMU-MOSI.}
\label{tab:weight_regression_mosi}
\setlength{\tabcolsep}{3.5pt} 
\begin{tabular}{l|cccc}
\noalign{\hrule height 1pt}
\textbf{Variable} & \textbf{Coefficient} & \textbf{Std. Err.} & \textbf{t-statistic} & \textbf{$\text{P}>|t|$} \\
\hline
Const & 1.8443 & 0.007 & 278.811 & $<$ 0.001 \\
Score Proximity & 1.743e-08 & 7.43e-09 & 2.345 & \textbf{0.019} \\
Conflict Count & 0.0040 & 0.004 & 0.913 & 0.361 \\
\noalign{\hrule height 1pt}
\multicolumn{5}{l}{N = 3125 (top 5\% hard-to-rank pairs) $|$ R-squared: 0.002} \\
\end{tabular}
\end{table}

\paragraph{(b) On MSD: Hardness is Overcoming the Text Bias Trap:} Sarcasm is defined by the incongruity between literal text and true intent. To prevent the tautological bias of extracting unimodal predictions from a jointly trained network, we trained independent unimodal experts and defined the Text Error Margin, which refers to the degree to which the independent text model makes a confident yet fundamentally incorrect prediction based on literal semantics. 

As shown in Table~\ref{tab:weight_regression_mustard}, an OLS regression reveals a highly significant positive relationship ($p < 0.001$) between the Text Error Margin and the penalty weights. Our analysis shows that when the text expert is misled by literal cues, GRCF identifies these samples as the hardest, assigning them the highest penalties. Rather than falling into the ``text bias trap'', GRCF leverages this stark contradiction to make predictions. Our Adaptive Error-weighting mechanism facilitates this by assigning high penalties early in training to samples where textual cues are deeply misleading, forcing the network to abandon literal interpretations and learn deep, non-linear multimodal contrast.

\begin{table}[t]
\small
\centering
\caption{OLS regression analyzing predictors of the penalty weights on MUStARD.}
\label{tab:weight_regression_mustard}
\setlength{\tabcolsep}{3.5pt} 
\begin{tabular}{l|cccc}
\noalign{\hrule height 1pt}
\textbf{Variable} & \textbf{Coefficient} & \textbf{Std. Err.} & \textbf{t-statistic} & \textbf{$\text{P}>|t|$} \\
\hline
Const & 18.3717 & 2.207 & 8.325 & $<$ 0.001 \\
Text Error Margin & 1.1976 & 0.243 & 4.933 & \textbf{$<$ 0.001} \\
\noalign{\hrule height 1pt}
\multicolumn{5}{l}{N = 539 $|$ R-squared: 0.043} \\
\end{tabular}
\end{table}

\paragraph{Summary:} These analyzes validate that the Error-weighting mechanism is contextually adaptive and accurately captures the underlying nature of distinct tasks. In MSA, it targets fine-grained sentiment intensities. In MSD, it navigates the complex semantics of sarcasm by prioritizing samples that require overcoming deep literal text biases, proving that the model learns the true essence of cross-modal interactions rather than relying on naive heuristics.

\subsection{Analysis of Dynamic Margin Mechanism}
To investigate how different grouping strategies for the Dynamic Margin affect performance, we conducted a comparative experiment on CH-SIMS v2, with results shown in Table~\ref{tab:bins_comparison}. We compare the ``Overlap, 5'' strategy (detailed in the Methodology~\ref{sec: overlap-group}) against three variants:
\begin{itemize}
    \item ``Overlap, 3'' and ``Overlap, 7'': Use 3 and 7 overlapping bins, respectively, to test granularity.
    \item ``Strict, 5'': Uses 5 non-overlapping bins (e.g., [-1.0, -0.6), [-0.6, -0.2), ...) to test the importance of the overlap itself.
\end{itemize}

Results in Table~\ref{tab:bins_comparison} validate that the ``Overlap, 5'' strategy significantly outperforms other configurations in almost all metrics (e.g., 74.3 Acc2). The performance drop of the ``Strict, 5'' strategy (73.2 Acc2) confirms that modeling semantic ambiguity via overlapping intervals is critical. 

Furthermore, the degraded results from ``Overlap, 3'' and ``Overlap, 7'' suggest that 5 bins provide the optimal granularity for capturing the sentiment spectrum. This phenomenon aligns with our proposed information efficiency criterion. Specifically, for CH-SIMS v2, the label span is 2.0 and the intrinsic annotation noise $\epsilon_y$ (proxied by the SOTA MAE) is approximately 0.27, yielding a theoretical resolution limit $K_{max} \approx 7$. When setting $K=7$, the theoretical group span $\Delta_7 \approx 0.28$ almost equates to the noise floor, causing the Dynamic Margin to overfit annotation ambiguity rather than learning valid semantic differences. Conversely, $K=5$ provides a span $\Delta_5 = 0.40$, which secures a sufficient margin above the noise $\epsilon_y$. Therefore, $K=5$ establishes the optimal signal-to-noise balance, acting as an optimal trade-off constrained by the physical semantic resolvability of the dataset.
\begin{table}[t]
\small
\centering
\caption{Comparison of different Dynamic Margin strategies on CH-SIMS v2.}
\label{tab:bins_comparison}
\small 
\setlength{\tabcolsep}{4pt}
\begin{tabular}{l|c|c|c|c|c|c} 
\noalign{\hrule height 1pt}
\textbf{Strategy} & \textbf{Acc5} $\uparrow$ & \textbf{Acc3} $\uparrow$ & \textbf{Acc2} $\uparrow$ & \textbf{F1-score} $\uparrow$ & \textbf{MAE} $\downarrow$ & \textbf{Corr} $\uparrow$ \\ \hline 
Overlap, 3 &  \textbf{54.6} & 73.8 & 81.2 & 81.2 & 0.307 & 0.689 \\
\textbf{Overlap, 5} & \underline{54.4} & \textbf{74.3} & \textbf{81.7} & \textbf{81.6} & \textbf{0.305} & \underline{0.691} \\
Overlap, 7 & 54.1 & \underline{74.1} & \underline{81.5} & \underline{81.5} & \underline{0.306} & \textbf{0.692} \\
Strict, 5  & 53.5 & 73.2 & 80.5 & 80.6 & 0.311 & 0.690\\
\noalign{\hrule height 1pt}
\end{tabular}
\end{table}

\subsection{Hyperparameter Stability Experiments}
\begin{figure}[t]
% \vspace{-0.2em}
\centering
\begin{subfigure}[b]{0.495\linewidth}
    \includegraphics[width=\linewidth]{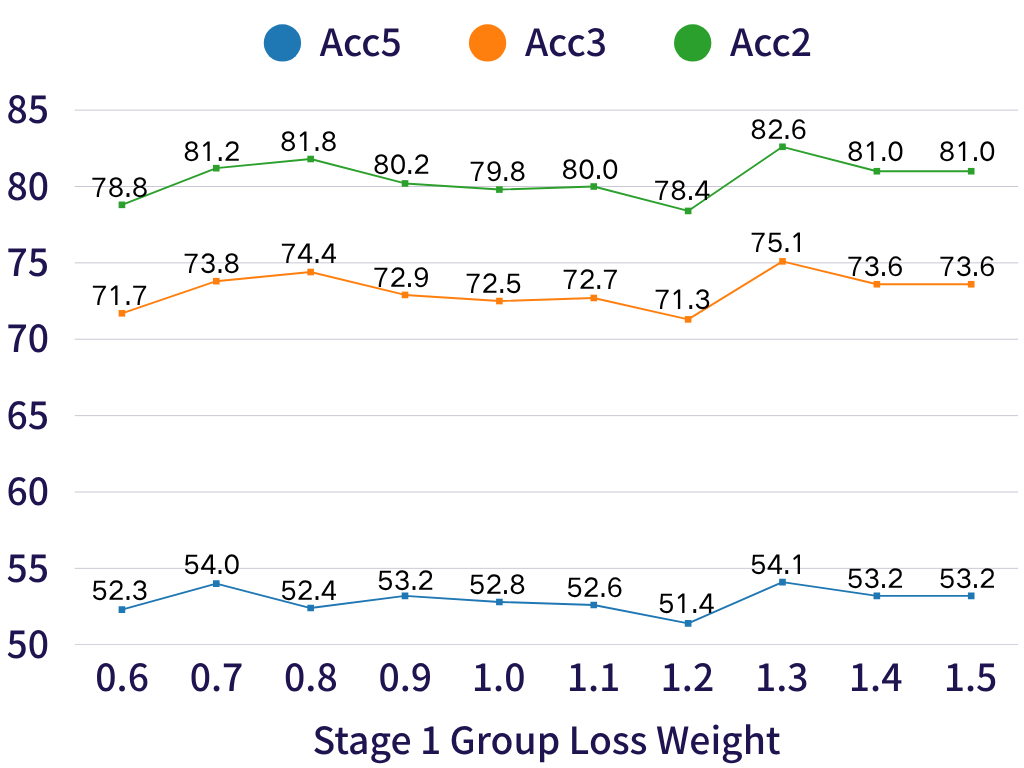}
    \label{fig:stage1_stability1}
\end{subfigure}
\hfill % 
\begin{subfigure}[b]{0.495\linewidth}
    \includegraphics[width=\linewidth]{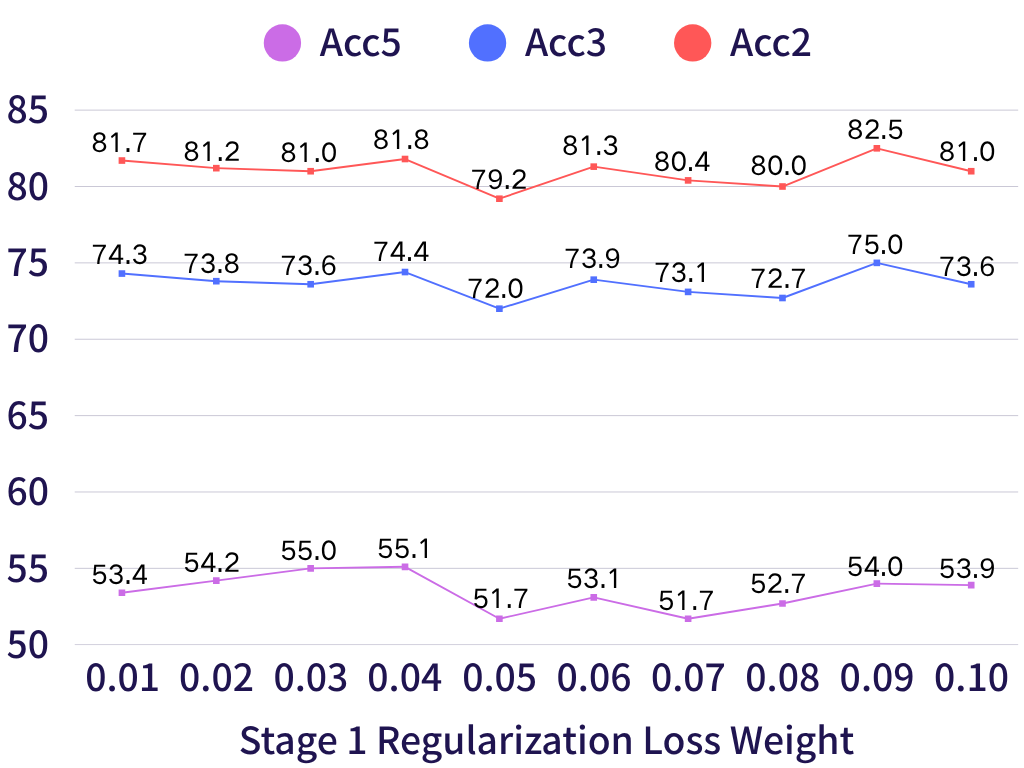}
    \label{fig:stage1_stability2}
\end{subfigure}
\caption{
    Performance with respect to the change of hyperparameters $\lambda_1$ and $\lambda_2$.
}
\label{fig:stage1_stability}
\end{figure}

In this section, we assess the impact of hyperparameters $\lambda_1$ and $\lambda_2$, which denote the weights of $\mathcal{L}_{\text{group}}$ and $\mathcal{L}_{\text{reg}}$ in Stage 1, respectively, on CH-SIMS v2. 

As shown in Figure~\ref{fig:stage1_stability}, GRCF delivers consistently satisfactory performance across a broad range of settings. For $\lambda_1$, the metrics remain remarkably stable as the value varies from 0.6 to 1.5. 
Notably, when $\lambda_1$ is set to 1.3, the model achieves an Acc2 of 82.6, which surpasses the result of 81.7 reported in Table 2. Similarly, for $\lambda_2$, performance fluctuations are marginal despite variations from 0.01 to 0.10. Interestingly, setting $\lambda_2$ to 0.04 yields a peak Acc5 of 55.1 and Acc3 of 74.4, again outperforming results in Table~\ref{tab:SIMSResult}.

This phenomenon demonstrates that GRCF is robust to the specific magnitude of loss weights. The potential of GRCF can be further unlocked through hyperparameter tuning.

\subsection{Computational Cost and Scalability}
While pair-wise ordinal learning introduces an $O(N^2)$ sample expansion during training, this computational overhead is confined to the training phase. During inference, GRCF reverts to a standard point-wise forward pass, processing each test sample independently without complex multi-branch interactions. 

As shown in Table~\ref{tab:comp_cost}, due to our streamlined unified encoder, GRCF's inference computational complexity (7.7 GFLOPs) is actually lower than that of the baseline model EMOE~\citep{Fang_2025_CVPR} and MOAC~\citep{mai2025learningbycomparing}. This reasonable one-time training cost yields a robust manifold for highly efficient, low-latency deployment.
\begin{table}[t]
\centering
\small
\caption{Comparison of computational cost with different baseline models on CMU-MOSI. Inference FLOPs in GFLOPs, Params in million.}
\label{tab:comp_cost}
\begin{tabular}{l|c|c|c} 
\noalign{\hrule height 1pt}
\textbf{Model} & \textbf{Format} & \textbf{Inference FLOPs} & \textbf{Params} \\ \hline 
EMOE~\citep{Fang_2025_CVPR} & Point-wise & 10.0 & 341.5 \\
% CaMIB~\citep{DBLP:journals/corr/abs-2509-21805} & Point-wise & 11.7 & 189.2 \\
MOAC~\citep{mai2025learningbycomparing} & Pair-wise & 10.2 & 186.9 \\
\hline
GRCF (Two-stage) & Pair-wise & 7.7 & 186.6 \\
\noalign{\hrule height 1pt}
\end{tabular}
\end{table}

\section{Conclusion and Future Work}
In this work, we introduced GRCF, which adapts the group-wise philosophy for ordinal representation learning in Affective Computing. Our core innovation is the Group-aware Ranking Loss, which builds an ordinally-consistent foundation by focusing on hard-to-rank pairs via an Error-weighting mechanism and enforcing semantically-proportional distances via a Dynamic Margin mechanism. GRCF achieves strong performance on regression benchmarks and demonstrates generalizability to classification tasks.

A key direction for future work is to address the current performance bottleneck in binary classification tasks. We attribute this limitation primarily to our existing multimodal fusion module; although it incorporates a gating mechanism, it remains relatively simple for modeling complex, heterogeneous cross-modal dynamics. Future research will focus on integrating more sophisticated fusion architectures to better capture nuanced affective cues and unresolved inter-modality conflicts.

\begin{acks}
This work is supported by the Guangdong Philosophy and Social Sciences Planning Project (No. GD26YJY34).
\end{acks}

\bibliographystyle{ACM-Reference-Format}
\bibliography{sample-base}

@String{Computing = "Computing" }

@String{Computer = "{IEEE} Computer" }

@article{Lan2019ALBERTAL,
  title={ALBERT: A Lite BERT for Self-supervised Learning of Language Representations},
  author={Zhenzhong Lan and Mingda Chen and Sebastian Goodman and Kevin Gimpel and Piyush Sharma and Radu Soricut},
  journal={ArXiv},
  year={2019},
  volume={abs/1909.11942},
  url={https://api.semanticscholar.org/CorpusID:202888986}
}

@inproceedings{devlin-etal-2019-bert,
    title = "{BERT}: Pre-training of Deep Bidirectional Transformers for Language Understanding",
    author = "Devlin, Jacob  and
      Chang, Ming-Wei  and
      Lee, Kenton  and
      Toutanova, Kristina",
    editor = "Burstein, Jill  and
      Doran, Christy  and
      Solorio, Thamar",
    booktitle = "Proceedings of the 2019 Conference of the North {A}merican Chapter of the Association for Computational Linguistics: Human Language Technologies, Volume 1 (Long and Short Papers)",
    month = jun,
    year = "2019",
    address = "Minneapolis, Minnesota",
    publisher = "Association for Computational Linguistics",
    url = "https://aclanthology.org/N19-1423/",
    doi = "10.18653/v1/N19-1423",
    pages = "4171--4186"
}

@inproceedings{
He2020DeBERTaDB,
title={DEBERTA: DECODING-ENHANCED BERT WITH DISENTANGLED ATTENTION},
author={Pengcheng He and Xiaodong Liu and Jianfeng Gao and Weizhu Chen},
booktitle={International Conference on Learning Representations},
year={2021},
url={https://openreview.net/forum?id=XPZIaotutsD}
}

@inproceedings{hazarika2020misamodalityinvariantspecificrepresentations,
  title={Misa: Modality-invariant and-specific representations for multimodal sentiment analysis},
  author={Hazarika, Devamanyu and Zimmermann, Roger and Poria, Soujanya},
  booktitle={Proceedings of the 28th ACM international conference on multimedia},
  pages={1122--1131},
  year={2020}
}

@inproceedings{wang2025dlf,
  title={DLF: Disentangled-language-focused multimodal sentiment analysis},
  author={Wang, Pan and Zhou, Qiang and Wu, Yawen and Chen, Tianlong and Hu, Jingtong},
  booktitle={Proceedings of the AAAI Conference on Artificial Intelligence},
  volume={39},
  number={20},
  pages={21180--21188},
  year={2025}
}

@inproceedings{burges2005learning,
  title={Learning to rank using gradient descent},
  author={Burges, Chris and Shaked, Tal and Renshaw, Erin and Lazier, Ari and Deeds, Matt and Hamilton, Nicole and Hullender, Greg},
  booktitle={Proceedings of the 22nd international conference on Machine learning},
  pages={89--96},
  year={2005}
}

@inproceedings{akiba2019optunanextgenerationhyperparameteroptimization,
  title={Optuna: A Next-generation Hyperparameter Optimization Framework},
  author={Akiba, Takuya and Sano, Shotaro and Yanase, Toshihiko and Ohta, Takeru and Koyama, Masanori},
  booktitle={KDD},
  year={2019}
}

@book{6287370,
  title={Advances in neural information processing systems 19: Proceedings of the 2006 conference},
  author={Sch{\"o}lkopf, Bernhard and Platt, John and Hofmann, Thomas},
  volume={19},
  year={2007},
  publisher={MIT press}
}

@inproceedings{mai2025learningbycomparing,
  title={Learning by Comparing: Boosting Multimodal Affective Computing through Ordinal Learning},
  author={Mai, Sijie and Zeng, Ying and Hu, Haifeng},
  booktitle={Proceedings of the ACM on Web Conference 2025},
  pages={2120--2134},
  year={2025}
}

@inproceedings{mmsa,
  title={Learning modality-specific representations with self-supervised multi-task learning for multimodal sentiment analysis},
  author={Yu, Wenmeng and Xu, Hua and Yuan, Ziqi and Wu, Jiele},
  booktitle={Proceedings of the AAAI conference on artificial intelligence},
  volume={35},
  number={12},
  pages={10790--10797},
  year={2021}
}

@article{HUANG2025102725,
title = {AtCAF: Attention-based causality-aware fusion network for multimodal sentiment analysis},
journal = {Information Fusion},
volume = {114},
pages = {102725},
year = {2025},
issn = {1566-2535},
doi = {https://doi.org/10.1016/j.inffus.2024.102725},
url = {https://www.sciencedirect.com/science/article/pii/S1566253524005037},
author = {Changqin Huang and Jili Chen and Qionghao Huang and Shijin Wang and Yaxin Tu and Xiaodi Huang}
}

@inproceedings{deva,
  title={Enriching Multimodal Sentiment Analysis through Textual Emotional Descriptions of Visual-Audio Content},
  author={Wu, Sheng and He, Dongxiao and Wang, Xiaobao and Wang, Longbiao and Dang, Jianwu},
  booktitle={Proceedings of the AAAI Conference on Artificial Intelligence},
  volume={39},
  number={2},
  pages={1601--1609},
  year={2025}
}

@article{MIB,
  title={Multimodal Information Bottleneck: Learning Minimal Sufficient Unimodal and Multimodal Representations},
  author={Mai, Sijie and Zeng, Ying and Hu, Haifeng},
  journal={IEEE Transactions on Multimedia},
  year={2023},
volume={25},
  pages={4121--4134},
  publisher={IEEE}
}

@inproceedings{ithp,
  title={Neuro-Inspired Information-Theoretic Hierarchical Perception for Multimodal Learning},
  author={Xiao, Xiongye and Liu, Gengshuo and Gupta, Gaurav and Cao, Defu and Li, Shixuan and Li, Yaxing and Fang, Tianqing and Cheng, Mingxi and Bogdan, Paul},
  booktitle={The Twelfth International Conference on Learning Representations},
year = {2024}
}

@ARTICLE{10224356,
  author={Mai, Sijie and Sun, Ya and Xiong, Aolin and Zeng, Ying and Hu, Haifeng},
  journal={IEEE Transactions on Multimedia}, 
  title={Multimodal Boosting: Addressing Noisy Modalities and Identifying Modality Contribution}, 
  year={2024},
  volume={26},
  number={},
  pages={3018-3033},
  doi={10.1109/TMM.2023.3306489}}

@article{DBLP:journals/corr/abs-2509-21805,
  author       = {Menghua Jiang and
                  Yuncheng Jiang and
                  Haifeng Hu and
                  Sijie Mai},
  title        = {Towards Minimal Causal Representations for Human Multimodal Language
                  Understanding},
  journal      = {CoRR},
  volume       = {abs/2509.21805},
  year         = {2025},
  url          = {https://doi.org/10.48550/arXiv.2509.21805},
  doi          = {10.48550/ARXIV.2509.21805},
  eprinttype    = {arXiv},
  eprint       = {2509.21805},
  bibsource    = {dblp computer science bibliography, https://dblp.org}
}

@article{tmson,
  title={Trustworthy multimodal fusion for sentiment analysis in ordinal sentiment space},
  author={Xie, Zhuyang and Yang, Yan and Wang, Jie and Liu, Xiaorong and Li, Xiaofan},
  journal={IEEE Transactions on Circuits and Systems for Video Technology},
  volume={34},
  number={8},
  pages={7657--7670},
  year={2024},
  publisher={IEEE}
}

@inproceedings{li2023decoupled,
  title={Decoupled multimodal distilling for emotion recognition},
  author={Li, Yong and Wang, Yuanzhi and Cui, Zhen},
  booktitle={Proceedings of the IEEE/CVF conference on computer vision and pattern recognition},
  pages={6631--6640},
  year={2023}
}

@InProceedings{Gao_2024_CVPR,
    author    = {Gao, Zixian and Jiang, Xun and Xu, Xing and Shen, Fumin and Li, Yujie and Shen, Heng Tao},
    title     = {Embracing Unimodal Aleatoric Uncertainty for Robust Multimodal Fusion},
    booktitle = {Proceedings of the IEEE/CVF Conference on Computer Vision and Pattern Recognition (CVPR)},
    month     = {June},
    year      = {2024},
    pages     = {26876-26885}
}

@InProceedings{Fang_2025_CVPR,
    author    = {Fang, Yiyang and Huang, Wenke and Wan, Guancheng and Su, Kehua and Ye, Mang},
    title     = {EMOE: Modality-Specific Enhanced Dynamic Emotion Experts},
    booktitle = {Proceedings of the Computer Vision and Pattern Recognition Conference (CVPR)},
    month     = {June},
    year      = {2025},
    pages     = {14314-14324}
}

@article{vaswani2017attention,
  title={Attention is all you need},
  author={Vaswani, Ashish and Shazeer, Noam and Parmar, Niki and Uszkoreit, Jakob and Jones, Llion and Gomez, Aidan N and Kaiser, {\L}ukasz and Polosukhin, Illia},
  journal={Advances in neural information processing systems},
  volume={30},
  year={2017}
}

@article{yannakakis2018ordinal,
  title={The ordinal nature of emotions: An emerging approach},
  author={Yannakakis, Georgios N and Cowie, Roddy and Busso, Carlos},
  journal={IEEE Transactions on Affective Computing},
  volume={12},
  number={1},
  pages={16--35},
  year={2018},
  publisher={IEEE}
}

@inproceedings{feng-etal-2024-knowledge,
    title = "Knowledge-Guided Dynamic Modality Attention Fusion Framework for Multimodal Sentiment Analysis",
    author = "Feng, Xinyu  and
      Lin, Yuming  and
      He, Lihua  and
      Li, You  and
      Chang, Liang  and
      Zhou, Ya",
    editor = "Al-Onaizan, Yaser  and
      Bansal, Mohit  and
      Chen, Yun-Nung",
    booktitle = "Findings of the Association for Computational Linguistics: EMNLP 2024",
    month = nov,
    year = "2024",
    address = "Miami, Florida, USA",
    publisher = "Association for Computational Linguistics",
    url = "https://aclanthology.org/2024.findings-emnlp.865/",
    doi = "10.18653/v1/2024.findings-emnlp.865",
    pages = "14755--14766"
}

@inproceedings{zadeh2018memory,
  title={Memory fusion network for multi-view sequential learning},
  author={Zadeh, Amir and Liang, Paul Pu and Mazumder, Navonil and Poria, Soujanya and Cambria, Erik and Morency, Louis-Philippe},
  booktitle={Proceedings of the AAAI conference on artificial intelligence},
  volume={32},
  number={1},
  year={2018}
}

@inproceedings{tsai2019multimodal,
  title={Multimodal transformer for unaligned multimodal language sequences},
  author={Tsai, Yao-Hung Hubert and Bai, Shaojie and Liang, Paul Pu and Kolter, J Zico and Morency, Louis-Philippe and Salakhutdinov, Ruslan},
  booktitle={Proceedings of the conference. Association for computational linguistics. Meeting},
  volume={2019},
  pages={6558},
  year={2019}
}

@inproceedings{luo2025towards,
  title={Towards explainable fusion and balanced learning in multimodal sentiment analysis},
  author={Luo, Miaosen and Jiang, Yuncheng and Mai, Sijie},
  booktitle={Proceedings of the 33rd ACM International Conference on Multimedia},
  pages={1997--2006},
  year={2025}
}

@inproceedings{rahman2020integrating,
  title={Integrating multimodal information in large pretrained transformers},
  author={Rahman, Wasifur and Hasan, Md Kamrul and Lee, Sangwu and Zadeh, Amir and Mao, Chengfeng and Morency, Louis-Philippe and Hoque, Ehsan},
  booktitle={Proceedings of the conference. Association for Computational Linguistics. Meeting},
  volume={2020},
  pages={2359},
  year={2020},
  organization={NIH Public Access}
}

@inproceedings{han-etal-2021-improving,
    title = "Improving Multimodal Fusion with Hierarchical Mutual Information Maximization for Multimodal Sentiment Analysis",
    author = "Han, Wei  and
      Chen, Hui  and
      Poria, Soujanya",
    editor = "Moens, Marie-Francine  and
      Huang, Xuanjing  and
      Specia, Lucia  and
      Yih, Scott Wen-tau",
    booktitle = "Proceedings of the 2021 Conference on Empirical Methods in Natural Language Processing",
    month = nov,
    year = "2021",
    address = "Online and Punta Cana, Dominican Republic",
    publisher = "Association for Computational Linguistics",
    url = "https://aclanthology.org/2021.emnlp-main.723/",
    doi = "10.18653/v1/2021.emnlp-main.723",
    pages = "9180--9192"
}

@inproceedings{zadeh2017tensor,
  title={Tensor Fusion Network for Multimodal Sentiment Analysis},
  author={Zadeh, Amir and Chen, Minghai and Poria, Soujanya and Cambria, Erik and Morency, Louis-Philippe},
  booktitle={Proceedings of the 2017 Conference on Empirical Methods in Natural Language Processing},
  pages={1103--1114},
  year={2017}
}

@inproceedings{zadeh2018multimodal,
  title={Multimodal language analysis in the wild: Cmu-mosei dataset and interpretable dynamic fusion graph},
  author={Zadeh, AmirAli Bagher and Liang, Paul Pu and Poria, Soujanya and Cambria, Erik and Morency, Louis-Philippe},
  booktitle={Proceedings of the 56th Annual Meeting of the Association for Computational Linguistics (Volume 1: Long Papers)},
  pages={2236--2246},
  year={2018}
}

@inproceedings{liu2022make,
  title={Make acoustic and visual cues matter: Ch-sims v2. 0 dataset and av-mixup consistent module},
  author={Liu, Yihe and Yuan, Ziqi and Mao, Huisheng and Liang, Zhiyun and Yang, Wanqiuyue and Qiu, Yuanzhe and Cheng, Tie and Li, Xiaoteng and Xu, Hua and Gao, Kai},
  booktitle={Proceedings of the 2022 international conference on multimodal interaction},
  pages={247--258},
  year={2022}
}

@inproceedings{liu2018efficient,
  title={Efficient low-rank multimodal fusion with modality-specific factors},
  author={Liu, Zhun and Shen, Ying and Lakshminarasimhan, Varun Bharadhwaj and Liang, Paul Pu and Zadeh, AmirAli Bagher and Morency, Louis-Philippe},
  booktitle={Proceedings of the 56th Annual Meeting of the Association for Computational Linguistics (Volume 1: Long Papers)},
  pages={2247--2256},
  year={2018}
}

@inproceedings{castro-etal-2019-towards,
    title = "Towards Multimodal Sarcasm Detection (An {\_}{O}bviously{\_} Perfect Paper)",
    author = "Castro, Santiago  and
      Hazarika, Devamanyu  and
      P{\'e}rez-Rosas, Ver{\'o}nica  and
      Zimmermann, Roger  and
      Mihalcea, Rada  and
      Poria, Soujanya",
    editor = "Korhonen, Anna  and
      Traum, David  and
      M{\`a}rquez, Llu{\'i}s",
    booktitle = "Proceedings of the 57th Annual Meeting of the Association for Computational Linguistics",
    month = jul,
    year = "2019",
    address = "Florence, Italy",
    publisher = "Association for Computational Linguistics",
    url = "https://aclanthology.org/P19-1455/",
    doi = "10.18653/v1/P19-1455",
    pages = "4619--4629"
}

@inproceedings{hasan-etal-2019-ur,
    title = "{UR}-{FUNNY}: A Multimodal Language Dataset for Understanding Humor",
    author = "Hasan, Md Kamrul  and
      Rahman, Wasifur  and
      Bagher Zadeh, AmirAli  and
      Zhong, Jianyuan  and
      Tanveer, Md Iftekhar  and
      Morency, Louis-Philippe  and
      Hoque, Mohammed (Ehsan)",
    editor = "Inui, Kentaro  and
      Jiang, Jing  and
      Ng, Vincent  and
      Wan, Xiaojun",
    booktitle = "Proceedings of the 2019 Conference on Empirical Methods in Natural Language Processing and the 9th International Joint Conference on Natural Language Processing (EMNLP-IJCNLP)",
    month = nov,
    year = "2019",
    address = "Hong Kong, China",
    publisher = "Association for Computational Linguistics",
    url = "https://aclanthology.org/D19-1211/",
    doi = "10.18653/v1/D19-1211",
    pages = "2046--2056"
}

@article{posner2005circumplex,
  title={The circumplex model of affect: An integrative approach to affective neuroscience, cognitive development, and psychopathology},
  author={Posner, Jonathan and Russell, James A and Peterson, Bradley S},
  journal={Development and psychopathology},
  volume={17},
  number={3},
  pages={715--734},
  year={2005},
  publisher={Cambridge University Press}
}

@article{zadeh2016multimodal,
  title={Multimodal sentiment intensity analysis in videos: Facial gestures and verbal messages},
  author={Zadeh, Amir and Zellers, Rowan and Pincus, Eli and Morency, Louis-Philippe},
  journal={IEEE Intelligent Systems},
  volume={31},
  number={6},
  pages={82--88},
  year={2016},
  publisher={IEEE}
}

@article{hou2019deep,
  title={Deep multimodal multilinear fusion with high-order polynomial pooling},
  author={Hou, Ming and Tang, Jiajia and Zhang, Jianhai and Kong, Wanzeng and Zhao, Qibin},
  journal={Advances in Neural Information Processing Systems},
  volume={32},
  year={2019}
}

@article{poria2017review,
  title={A review of affective computing: From unimodal analysis to multimodal fusion},
  author={Poria, Soujanya and Cambria, Erik and Bajpai, Rajiv and Hussain, Amir},
  journal={Information fusion},
  volume={37},
  pages={98--125},
  year={2017},
  publisher={Elsevier}
}

@inproceedings{sun2022cubemlp,
  title={CubeMLP: An MLP-based model for multimodal sentiment analysis and depression estimation},
  author={Sun, Hao and Wang, Hongyi and Liu, Jiaqing and Chen, Yen-Wei and Lin, Lanfen},
  booktitle={Proceedings of the 30th ACM international conference on multimedia},
  pages={3722--3729},
  year={2022}
}

@article{zhu2023multimodal,
  title={Multimodal sentiment analysis based on fusion methods: A survey},
  author={Zhu, Linan and Zhu, Zhechao and Zhang, Chenwei and Xu, Yifei and Kong, Xiangjie},
  journal={Information Fusion},
  volume={95},
  pages={306--325},
  year={2023},
  publisher={Elsevier}
}

@inproceedings{liang2019learning,
    title = "Learning Representations from Imperfect Time Series Data via Tensor Rank Regularization",
    author = "Liang, Paul Pu  and
      Liu, Zhun  and
      Tsai, Yao-Hung Hubert  and
      Zhao, Qibin  and
      Salakhutdinov, Ruslan  and
      Morency, Louis-Philippe",
    editor = "Korhonen, Anna  and
      Traum, David  and
      M{\`a}rquez, Llu{\'i}s",
    booktitle = "Proceedings of the 57th Annual Meeting of the Association for Computational Linguistics",
    month = jul,
    year = "2019",
    address = "Florence, Italy",
    publisher = "Association for Computational Linguistics",
    url = "https://aclanthology.org/P19-1152/",
    doi = "10.18653/v1/P19-1152",
    pages = "1569--1576"
}

@article{li2023interpretable,
  title={Interpretable multimodal sentiment analysis based on textual modality descriptions by using large-scale language models},
  author={Li, Sixia and Okada, Shogo},
  journal={arXiv preprint arXiv:2305.06162},
  year={2023}
}

@inproceedings{stoehr2023sentiment,
  title={Sentiment as an ordinal latent variable},
  author={Stoehr, Niklas and Cotterell, Ryan and Schein, Aaron},
  booktitle={Proceedings of the 17th Conference of the European Chapter of the Association for Computational Linguistics},
  pages={103--115},
  year={2023}
}

@inproceedings{cheng2021multimodal,
  title={Multimodal phased transformer for sentiment analysis},
  author={Cheng, Junyan and Fostiropoulos, Iordanis and Boehm, Barry and Soleymani, Mohammad},
  booktitle={Proceedings of the 2021 Conference on Empirical Methods in Natural Language Processing},
  pages={2447--2458},
  year={2021}
}

@misc{meng2026trisubspacesdisentanglementmultimodalsentiment,
      title={Tri-Subspaces Disentanglement for Multimodal Sentiment Analysis}, 
      author={Chunlei Meng and Jiabin Luo and Zhenglin Yan and Zhenyu Yu and Rong Fu and Zhongxue Gan and Chun Ouyang},
      year={2026},
      eprint={2602.19585},
      archivePrefix={arXiv},
      primaryClass={cs.MM},
      url={https://arxiv.org/abs/2602.19585}, 
}

@misc{qian2025decalignhierarchicalcrossmodalalignment,
      title={DecAlign: Hierarchical Cross-Modal Alignment for Decoupled Multimodal Representation Learning}, 
      author={Chengxuan Qian and Shuo Xing and Shawn Li and Yue Zhao and Zhengzhong Tu},
      year={2025},
      eprint={2503.11892},
      archivePrefix={arXiv},
      primaryClass={cs.CV},
      url={https://arxiv.org/abs/2503.11892}, 
}

@article{thurstone2017law,
  title={A law of comparative judgment.},
  author={Thurstone, Louis L},
  journal={Psychological review},
  volume={34},
  number={4},
  pages={273},
  year={1927},
  publisher={Psychological Review Company}
}

@ARTICLE{11493959,
  author={Mai, Sijie and Lin, Ronghao and Hu, Haifeng},
  journal={IEEE Transactions on Multimedia}, 
  title={An Ordinal Learning Framework with Sentiment Decouple for Multimodal Sentiment Analysis}, 
  year={2026},
  volume={},
  number={},
  pages={1-16},
  doi={10.1109/TMM.2026.3687019}}

@article{friston2010free,
  title={The free-energy principle: a unified brain theory?},
  author={Friston, Karl},
  journal={Nature reviews neuroscience},
  volume={11},
  number={2},
  pages={127--138},
  year={2010},
  publisher={Nature publishing group}
}

@article{russell1980circumplex,
  title={A circumplex model of affect.},
  author={Russell, James A},
  journal={Journal of personality and social psychology},
  volume={39},
  number={6},
  pages={1161},
  year={1980},
  publisher={American Psychological Association}
}

@article{cowen2017self,
  title={Self-report captures 27 distinct categories of emotion bridged by continuous gradients},
  author={Cowen, Alan S and Keltner, Dacher},
  journal={Proceedings of the national academy of sciences},
  volume={114},
  number={38},
  pages={E7900--E7909},
  year={2017},
  publisher={National Academy of Sciences}
}

@article{Ekman01051992,
author = {Paul Ekman},
title = {An argument for basic emotions},
journal = {Cognition and Emotion},
volume = {6},
number = {3-4},
pages = {169--200},
year = {1992},
publisher = {Routledge},
doi = {10.1080/02699939208411068},
URL = { 
        https://doi.org/10.1080/02699939208411068
},
eprint = { 
        https://doi.org/10.1080/02699939208411068 
}
}
\end{document}